\documentclass[preprint,12pt]{elsarticle}

\usepackage{amssymb}
\usepackage{amsmath}
\usepackage{lipsum}
\usepackage{booktabs}
\usepackage{multirow}
\usepackage[hidelinks]{hyperref}
\usepackage{float}
\usepackage{xcolor}
\newcommand{\rev}[1]{#1}

\newcommand{\revNew}[1]{#1}

\journal{Neurocomputing}

\begin{document}

\begin{frontmatter}

%% Title, authors and addresses

%% use the tnoteref command within \title for footnotes;
%% use the tnotetext command for theassociated footnote;
%% use the fnref command within \author or \affiliation for footnotes;
%% use the fntext command for theassociated footnote;
%% use the corref command within \author for corresponding author footnotes;
%% use the cortext command for theassociated footnote;
%% use the ead command for the email address,
%% and the form \ead[url] for the home page:
%% \title{Title\tnoteref{label1}}
%% \tnotetext[label1]{}
%% \author{Name\corref{cor1}\fnref{label2}}
%% \ead{email address}
%% \ead[url]{home page}
%% \fntext[label2]{}
%% \cortext[cor1]{}
%% \affiliation{organization={},
%%             addressline={},
%%             city={},
%%             postcode={},
%%             state={},
%%             country={}}
%% \fntext[label3]{}

\title{CMHANet: A Cross-Modal Hybrid Attention Network for Point Cloud Registration} %% Article title

%% use optional labels to link authors explicitly to addresses:
%% \author[label1,label2]{}
%% \affiliation[label1]{organization={},
%%             addressline={},
%%             city={},
%%             postcode={},
%%             state={},
%%             country={}}
%%
%% \affiliation[label2]{organization={},
%%             addressline={},
%%             city={},
%%             postcode={},
%%             state={},
%%             country={}}

\address[1]{School of Software, Xi’an Jiaotong University, No. 28 Xianning West Road, China}
\address[2]{CEPRI, China Electric Power Research Institute
No. 15 Xiaoying East Road, Qinghe, Haidian District, Beijing, China}

\author[1]{Dongxu Zhang}
\author[2]{Yingsen Wang} 
\author[1]{Yiding Sun}
\author[1]{Haoran Xu}
\author[1]{Peilin Fan}
\author[1]{Jihua Zhu\corref{cor1}}

\cortext[cor1]{Corresponding author}

%% Abstract
\begin{abstract}
%% Text of abstract
Robust point cloud registration is a fundamental task in 3D computer vision and geometric deep learning, essential for applications such as large-scale 3D reconstruction, augmented reality, and scene understanding. However, the performance of established learning-based methods often degrades in complex, real world scenarios characterized by incomplete data, sensor noise, and low overlap regions.
To address these limitations, we propose CMHANet, a novel Cross-Modal Hybrid Attention Network. Our method integrates the fusion of rich contextual information from 2D images with the geometric detail of 3D point clouds, yielding a comprehensive and resilient feature representation. Furthermore, we introduce an innovative optimization function based on contrastive learning, which enforces geometric consistency and significantly improves the model's robustness to noise and partial observations.
We evaluated CMHANet on the 3DMatch and the challenging 3DLoMatch datasets. \rev{Additionally, zero-shot evaluations on the TUM RGB-D SLAM dataset verify the model's generalization capability to unseen domains.} The experimental results demonstrate that our method achieves substantial improvements in both registration accuracy and overall robustness, outperforming current techniques. We also release our code in \href{https://github.com/DongXu-Zhang/CMHANet}{https://github.com/DongXu-Zhang/CMHANet}.

\end{abstract}

%%Graphical abstract
% \begin{graphicalabstract}
% %\includegraphics{grabs}
% \end{graphicalabstract}

% %%Research highlights
% \begin{highlights}
% \item Research highlight 1
% \item Research highlight 2
% \end{highlights}

%% Keywords
\begin{keyword}
%% keywords here, in the form: keyword \sep keyword
Point cloud registration \sep Point matching \sep Deep learning for visual perception
%% PACS codes here, in the form: \PACS code \sep code

%% MSC codes here, in the form: \MSC code \sep code
%% or \MSC[2008] code \sep code (2000 is the default)

\end{keyword}

\end{frontmatter}

%% Add \usepackage{lineno} before \begin{document} and uncomment 
%% following line to enable line numbers
%% \linenumbers

%% main text
%%
\section{INTRODUCTION}
\label{introduction}

Point cloud registration, the process of aligning two or more 3D point sets into a unified coordinate system, is a fundamental problem in 3D computer vision and geometric deep learning \citep{zhang2024comprehensive,wang2022efficient,zhang2026igasa}. This task is essential for a wide range of applications, including 3D scene reconstruction \citep{shi2023iteration,li2024dbdnet}, augmented reality, and object shape analysis \citep{wang2022residual,huang2024kdd}. Despite its importance, achieving robust registration remains a significant challenge, particularly when dealing with real-world data affected by sensor noise, sparsity, and irregular sampling \citep{bai2021pointdsc}. Traditional methods \citep{jost2002fast,vizzo2023kiss,sharp2002icp}, while foundational, primarily rely on geometric properties and often overlook valuable contextual information, such as texture and semantic cues available in 2D images.
   \begin{figure}[t]
      \centering
      \includegraphics[width=0.85\linewidth]{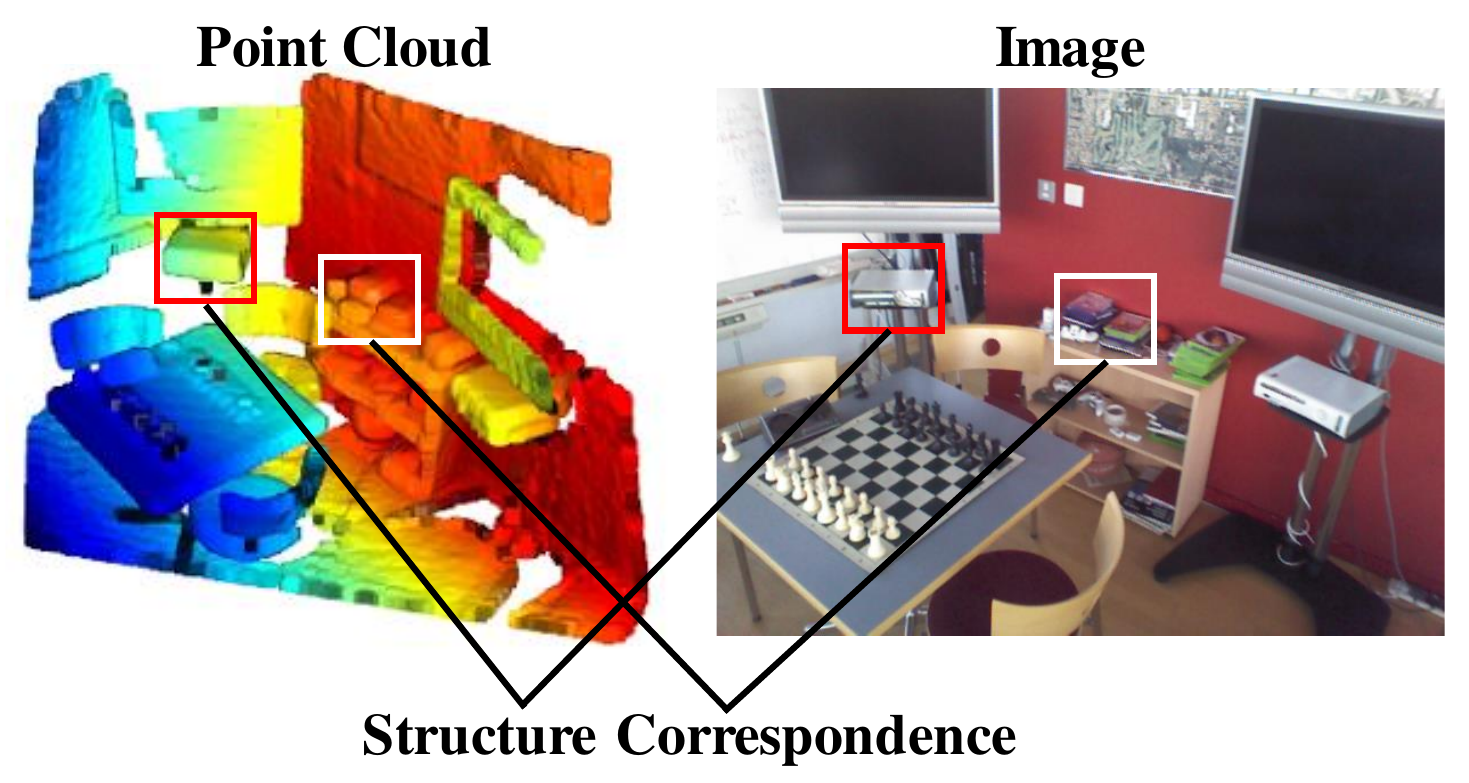}
      \caption{The pixel-to-point correspondence between the point cloud and the image has been established by external parameter calibration, so this method is reasonable, and can effectively combine the global texture features of the image with the local geometric characteristics of the point cloud.}
      \label{figurelabel}
   \end{figure}
   
The advent of deep learning has introduced a new paradigm for point cloud analysis \citep{jiang2025hybrid,sun2025hyperpoint,sun2026alignadaptrethinkingparameterefficient,sun6064487curve3d}, enabling end-to-end neural networks to learn powerful feature representations directly from raw data \citep{huang2021predator,yu2021cofinet,han2025rethinking}. Within the neural computing domain, network architectures have evolved to better capture the complex structures inherent in point clouds. While Convolutional Neural Networks (CNNs) are effective at extracting local features, their constrained receptive fields can limit their ability to model long-range dependencies. In contrast, transformer-based architectures \citep{qin2022geometric,yang2022one,yu2023rotation,chen2023sira,zhang2026pointcot} have demonstrated a strong capacity for capturing global context, a crucial characteristic for understanding the overall structure of a scene. Furthermore, insights from related fields like deformable image registration have highlighted the advantages of incorporating hierarchical and attention mechanisms to model multi-scale relationships \citep{xie2023cross,xu2024igreg}.

   \begin{figure}[t]
      \centering
      \includegraphics[width=0.7\linewidth]{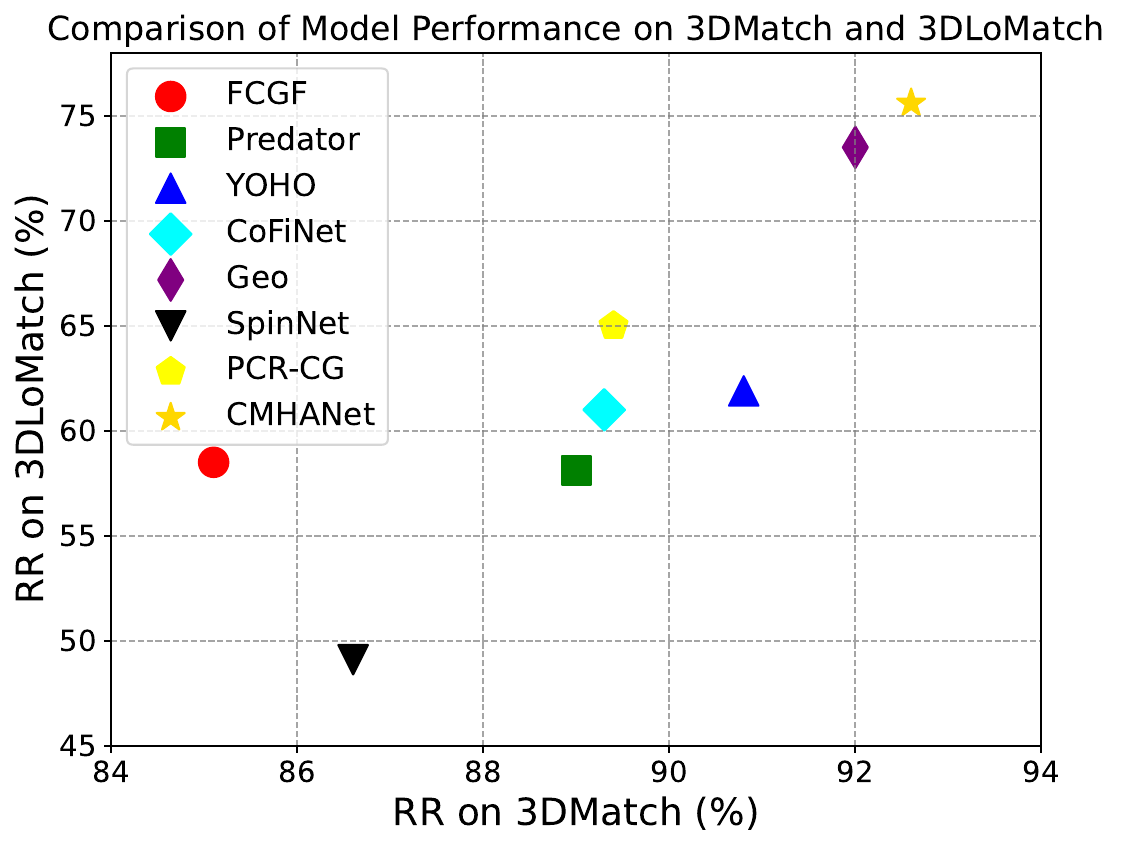}
      \caption{\rev{The Registration Recall (RR) is plotted on the x-axis for 3DMatch and on
the y-axis for 3DLoMatch. CMHANet stands out by consistently achieving the
highest RR.}}
      \label{figurelabel2}
   \end{figure}

A key opportunity to advance these learning-based methods lies in leveraging multimodal data. Sensor suites that pair depth sensors with RGB cameras are now commonplace, providing complementary data streams \citep{zeng20173dmatch}. Although point clouds encode precise 3D geometry, they often lack descriptive texture. Conversely, 2D images provide dense, texture-rich, and semantic context but lack explicit 3D information \citep{xu2024igreg}. As illustrated in Figure.~\ref{figurelabel}, fusing these complementary data sources allows for a more comprehensive scene understanding, thereby improving correspondence matching and registration robustness.

Motivated by these insights, we propose CMHANet: a Cross-Modal Hybrid Attention Network for point cloud registration. Our method is architected to effectively fuse these distinct data modalities. It utilizes separate feature encoders for the point cloud and image data, followed by a multi-stage cross-modal attention pipeline. This design allows CMHANet to build a richer joint feature space and introduces a hybrid attention mechanism that refines the interaction between geometric and visual features for more accurate superpoint matching.

The contributions of this paper are summarized as follows:
\begin{itemize}
\item We present a novel network architecture that seamlessly integrates 3D geometric and 2D texture information, generating a more discriminative feature representation for point cloud registration.
\item We develop a hybrid attention mechanism designed to intelligently model the interplay between 2D and 3D features, enabling precise and adaptive multimodal correspondence matching.
\item We formulate a detailed optimization objective that jointly promotes geometric fidelity and semantic coherence across the disparate data streams.
\end{itemize}

We validated the performance of CMHANet on the demanding 3DMatch \citep{zeng20173dmatch} and 3DLoMatch \citep{huang2021predator} benchmarks. Our empirical findings demonstrate that CMHANet consistently outperforms contemporary methods, as shown by its leading Registration Recall metrics across various overlap conditions (Figure.~\ref{figurelabel2}).
\section{RELATED WORK}
%%\label{}
\subsection{Correspondence-based Point Cloud Registration Methods}

Point Cloud Registration (PCR), which seeks to align multiple 3D point sets into a common coordinate system by estimating a rigid transformation \citep{hou2021pri3d}, stands as a foundational challenge in 3D computer vision. Its successful execution is a prerequisite for numerous applications, including large-scale 3D reconstruction and augmented reality \citep{bai2021pointdsc,el2021unsupervisedr}. The landscape of PCR methodologies can be broadly categorized into two principal paradigms: optimization-centric and correspondence-based.

The optimization-centric paradigm has been historically dominated by the Iterative Closest Point (ICP) algorithm \citep{sharp2002icp}. While foundational, the standard ICP algorithm is sensitive to the initial alignment and can be computationally intensive, often converging to local minima. Consequently, a variety of extensions have been developed, such as Fast-ICP \citep{jost2002fast}, Go-ICP \citep{yang2015go}, and the more recent Kiss-ICP \citep{vizzo2023kiss}, each aiming to improve the algorithm's robustness, efficiency, or convergence properties. However, as these methods primarily perform local refinement, their effectiveness is limited in scenarios with significant initial misalignments or low overlap.

In parallel, correspondence-based techniques have advanced significantly, largely driven by the adoption of deep learning \citep{rusu2009fast,gojcic2019perfect}. This category has evolved from traditional methods using handcrafted local descriptors based on geometric or statistical properties \citep{rusu2008aligning,choy2019fully} to the use of sophisticated neural network architectures. A development has been the integration of attention mechanisms, which have proven effective for capturing both global context and fine-grained local geometric details within point clouds \citep{qin2022geometric,yang2021focal}. More recently, the field has progressed towards end-to-end learning architectures that formulate the entire registration pipeline as a single optimization problem, allowing for the learning of transformations from raw data \citep{xie2024hecpg}.

Despite these advances, the majority of deep learning-based methods still focus exclusively on geometric information, overlooking the rich, complementary data available in associated 2D imagery. This unimodal method can limit feature discriminability, particularly in texture-rich environments or scenarios where geometric cues are ambiguous. Our work addresses this critical gap by proposing a novel correspondence-based framework that explicitly integrates 2D visual features with 3D geometric structures, creating a more robust and descriptive representation for accurate point cloud registration.

\subsection{Multimodal Fusion and Attention Mechanisms}

A promising direction for creating more discriminative feature representations is multimodal fusion~\citep{gong2025med,ijcv,zhang2026chain}, which integrates the sparse geometric structure of 3D point clouds with the dense textural and semantic context of 2D images \citep{xu2024igreg,zeng20173dmatch}. Several recent works~\citep{10298249tnnls,du2023degradationtgrs,che2025lemon,che2025stitch} have explored this concept. For example, IMFNet \citep{huang2022imfnet} employs attention mechanisms for feature fusion, while CMIGNet \citep{xie2023cross} leverages specialized convolutions such as KPConv \citep{thomas2019kpconv} to combine 2D semantic information with 3D geometry.

Central to many effective fusion strategies is the attention mechanism~\citep{chen2025uni,lan2026performance,lan2026reco,xue2024integrating}, particularly as implemented within Transformer architectures \citep{agarwal2023attention}. Unlike convolutional layers, which are constrained to local receptive fields, attention allows models such as GeoTransformer \citep{qin2022geometric} and CMDGAT \citep{guo2024learning} to model long-range dependencies. This capacity to capture both global and local relationships is critical for learning robust inter- and intra-modal correspondences for registration~\citep{zhang2026notallqueries,zhang2025ascot}.

However, while these methods confirm the potential of multimodal fusion, they often rely on generic fusion mechanisms. The design of sophisticated neural architectures that can intelligently and adaptively model the interplay between geometric and visual features remains a key challenge. Our work directly addresses this by introducing a hybrid-attention network designed for more precise and context-aware multimodal fusion.

\subsection{Robustness Challenges in Point Cloud Registration}

The practical performance of point cloud registration models is often limited by imperfections in real-world sensor data \citep{jiang2025hybrid,guo2024learning}. Data from sources like LiDAR and RGB-D cameras is frequently corrupted by noise, sparsity, occlusions, and non-uniform sampling, especially on challenging textureless or reflective surfaces \citep{huang2024kdd,wang2022you,li2023synergy}. From a neural computing perspective, these data imperfections introduce a cascade of challenges for deep learning models. They degrade the quality of learned features, leading to ambiguous and less discriminative representations that hinder the establishment of reliable correspondences. This problem is particularly acute in scenes with low overlap or repetitive geometric structures. Ultimately, these ambiguous correspondences create a difficult optimization landscape, making models susceptible to converging on incorrect alignments. The demand for computational efficiency further complicates this issue, often forcing a trade--off between model complexity and the registration accuracy required for downstream tasks.

\section{METHODS}
\begin{figure}[t]
      \centering

      \includegraphics[width=1\linewidth]{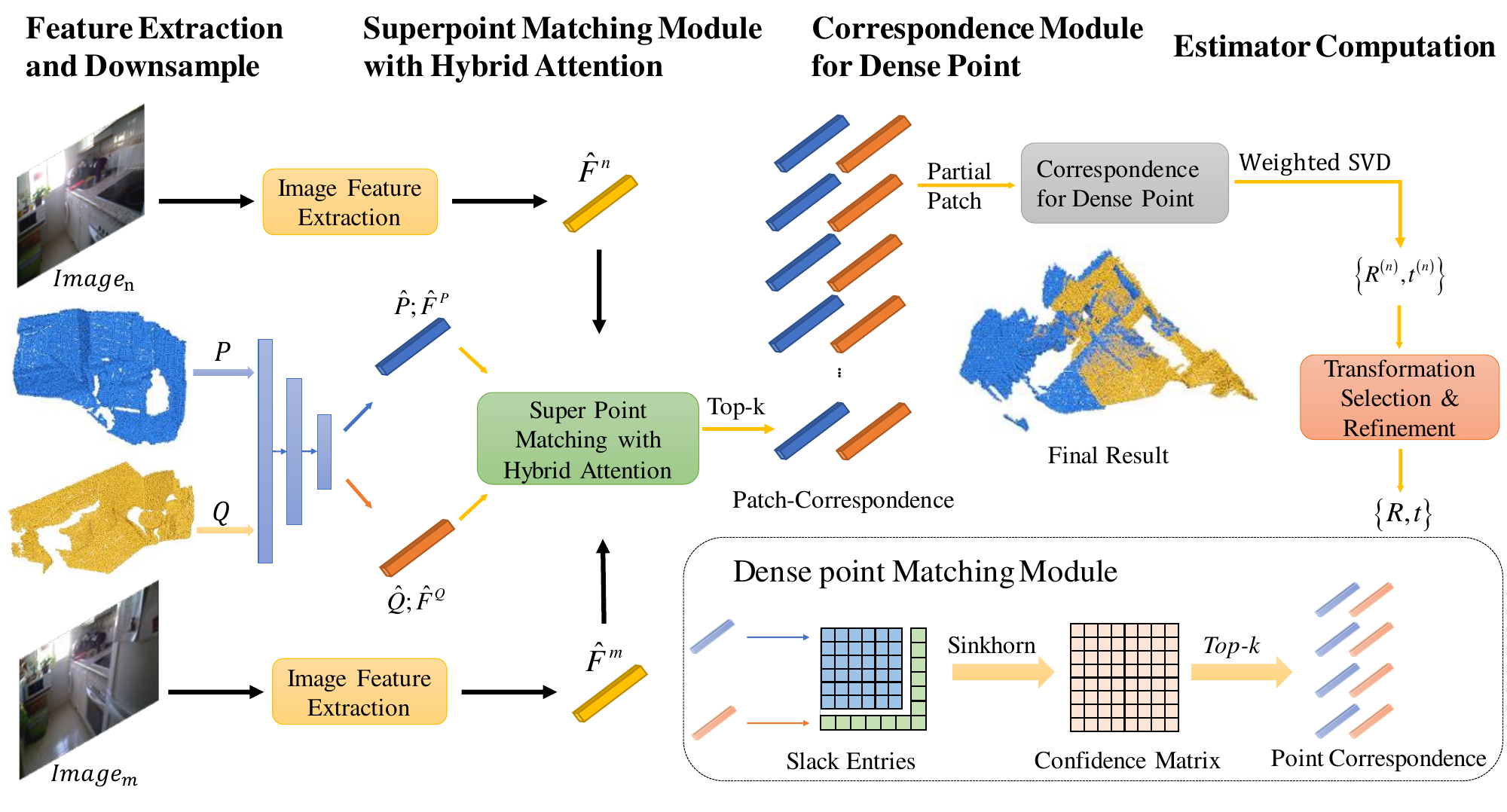}
      \caption{The workflow overview of CMHANet. Our method processes raw point clouds and images through feature extraction, employs a hybrid-attention mechanism for superpoint matching, refines to dense point correspondences, and finally computes the rigid transformation for alignment.}
      \label{figurelabe2}
   \end{figure}

Our method, as illustrated in Figure.~\ref{figurelabe2}, performs multimodal point cloud registration through a multi-stage architecture designed to fuse 3D geometric data and 2D visual information. The pipeline consists of the following interconnected modules.

First, the Feature Extraction and Downsampling module processes the raw inputs. It uses two parallel encoders: a point cloud encoder and an image encoder. The point cloud encoder extracts key geometric features from the 3D data while simultaneously downsampling it to produce a sparse set of representative grouped keypoints, referred to as superpoints. Concurrently, the image encoder extracts relevant visual features from the corresponding 2D image.

Subsequently, the outputs are then processed by the Superpoint Matching Module with Hybrid Attention, which serves as the core of our cross-modal fusion strategy. This module is equipped with three distinct attention mechanisms that explicitly model the complex relationships between the 3D geometric features and the 2D visual features. By learning these cross-modal dependencies, the network moves beyond simple feature concatenation to establish more robust initial correspondences between the superpoints.

Next, the Dense Correspondence Module refines these coarse-level associations. Using the established superpoint matches as a guide, this module infers dense point-to-point correspondences across the original, full-resolution point clouds. This step leverages the context captured in the previous stage to resolve local ambiguities and establish a detailed alignment relationship map.

Finally, the Transformation Estimation module computes the output transformation. It takes the dense point correspondences as input and calculates the rigid transformation (rotation and translation) that optimally aligns the source point cloud with the target point cloud.

\subsection{Problem Statement} 
We define the source point cloud and the target point cloud as $P$ and $Q$, respectively. \rev{The main objective of registration is to find a rigid transformation $T=(R, t)$ that aligns $P$ with $Q$. And this transformation consists of a rotation matrix $R\in SO(3)$ and a translation vector $t\in R^3$.}

\subsection{Feature Extraction and Downsample}
The initial stage of our network extracts features from both the 3D point cloud and 2D image modalities, while also downsampling the point cloud to generate a sparse set of representative keypoints, which we term superpoints.

%\subsubsection{Point cloud feature extraction}
\textbf{Point Cloud Feature Extraction}\quad \rev{For the input point clouds $P$ and $Q$, we first perform feature extraction and downsampling using the Kernel Point Convolution with Feature Pyramid Network (KPConv-FPN) backbone~\cite{thomas2019kpconv}. The final layer downsampled points (namely superpoints) are represented as $ S^P = \left\{ s^i_p \in \mathbb{R}^{N_P \times 3} \right\} \quad \text{and} \quad S^Q = \left\{ s^j_q \in \mathbb{R}^{N_Q \times 3} \right\}$.
The corresponding features are denoted as $F^p_s \in \mathbb{R}^{N_P \times d} \quad \text{and} \quad F^q_s \in \mathbb{R}^{N_Q \times d}$.
To link the original dense points to these superpoints, we apply Nearest-Superpoint Aggregation for Point Feature Grouping.} For each point $p_i \in P$ and $q_j \in Q$, we associate it with the nearest superpoint. Next, we define the distance from point $p_i$ to its nearest superpoint $\mathit{s}_k^*$ as $d(p_i, s_k) = \| p_i - s_k \|_2$,
using the Euclidean distance. Each point is then aggregated to its corresponding superpoint as:
\begin{equation}
G_{s_k} = \left\{ p_i \in P \mid \arg \min_{k} \left( d(p_i, s_k) \right) = k \right\}.
\end{equation}

The features of these points are also associated with the feature matrix of the superpoints.

\begin{figure*}[t]
  \centering
  \includegraphics[width=1\linewidth]{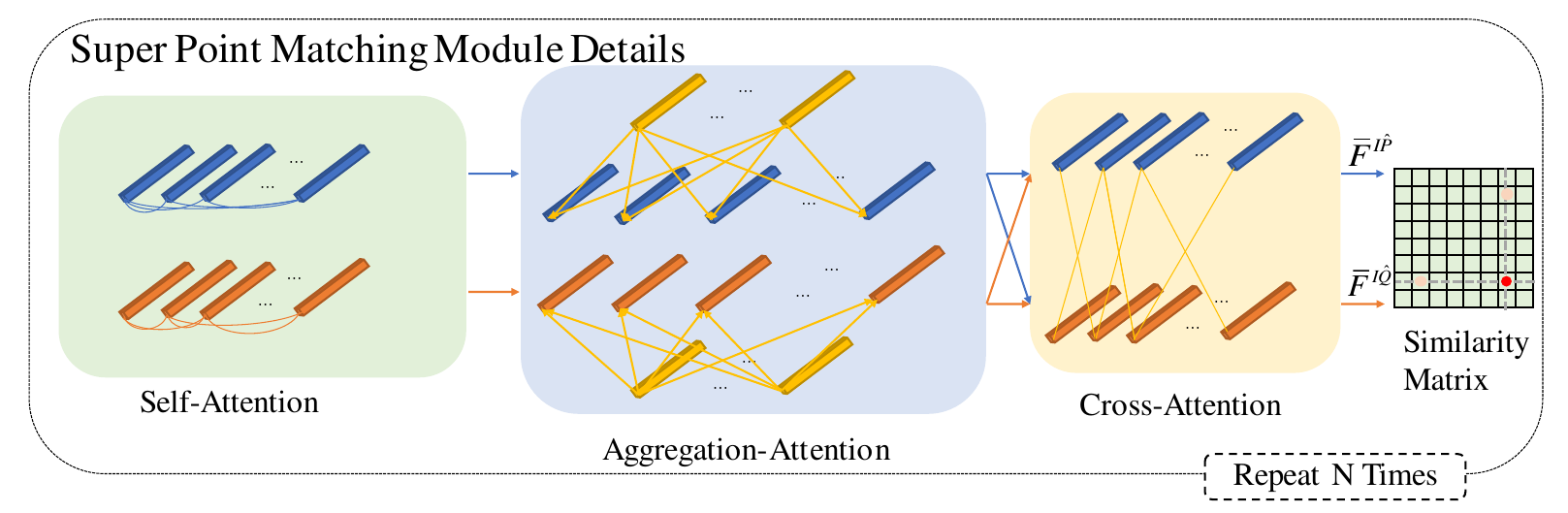}
  \caption{
  Details of the superpoint matching module. This module employs a multi-stage attention mechanism: Self-Attention refines individual modality features, followed by Aggregation-Attention for structured intra-modal feature integration, and finally Cross-Attention to fuse features across modalities, ultimately generating a similarity matrix for robust superpoint matching. \revNew{For visualization, the blue blocks represent the source point cloud features ($\hat{F}^P$), the orange blocks represent the target point cloud features ($\hat{F}^Q$), and the yellow lines indicate the attention interactions.}
  }
  \label{figurelabe3}
\end{figure*}

%\subsubsection{Image feature extraction}
\textbf{Image Feature Extraction}\quad We use the ResUNet-50 backbone network to extract image features. For the images of the source and target point clouds $\textit{Image}_n, \textit{Image}_m $, corresponding features are denoted as $\hat{\mathit{F}}^n, \hat{\mathit{F}}^m $.

\subsection{Superpoint Matching Module with Hybrid Attention}

Global contextual information plays a crucial role in many 3D vision tasks. Therefore, we propose a novel attention mechanism, referred to as Hybrid Attention, which is designed to efficiently aggregate and enhance point cloud feature representations. \rev{Our Hybrid Attention consists of three distinct attention modules: the geometric self-attention module, which learns the features of the point cloud itself; the aggregation attention module, which fuses 2D image features into the corresponding point cloud information; and the geometric cross-attention module, which learns the consistency features between the source and target point clouds.}

As shown in Figure.~\ref{figurelabe3}, the three modules alternate for $N$ iterations to extract hybrid features for reliable superpoint matching. This iterative process progressively refines the feature representations, producing robust hybrid features that are crucial for reliable superpoint matching and ultimately improve the overall performance of point cloud registration.

\begin{figure}[t]
  \centering
  \includegraphics[width=1\linewidth]{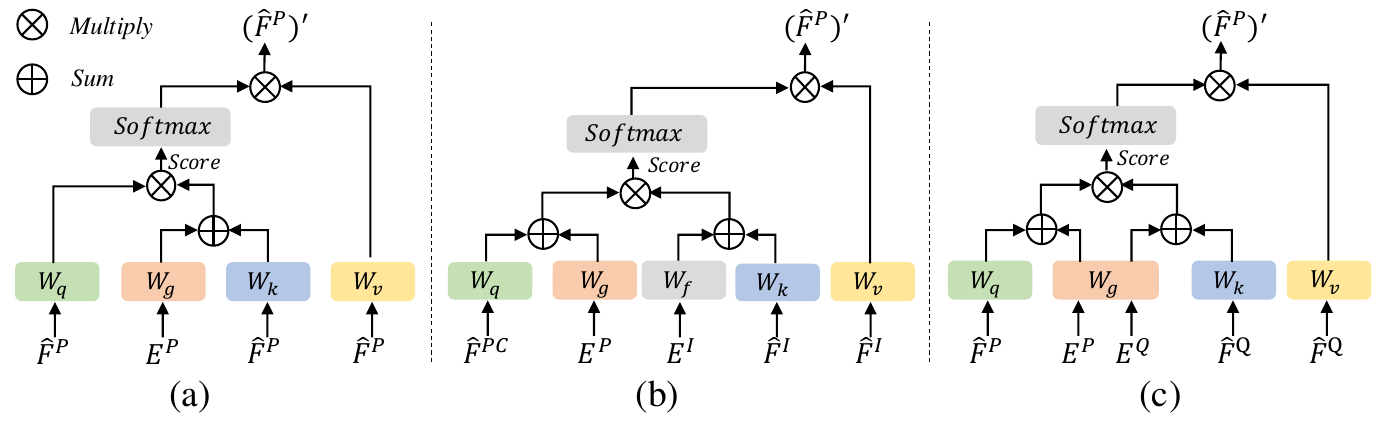}
  \caption{Computation graph of the Hybrid Attention mechanism. \rev{In the diagrams, the label $F$ denotes feature representations ($e.g.,$ \revNew{$\hat{F}^{PC}$ for generic point cloud features, $\hat{F}^P$ and $\hat{F}^Q$ for source and target point cloud features (in a, c),} $\hat{F}^I$ for image features), and $E$ represents geometric positional embeddings. The three modules serve distinct roles. (a) Geometric Self-Attention captures global structural relationships within a single point cloud; (b) Geometric Aggregation-Attention fuses 2D visual context from images into 3D geometric features; and (c) Geometric Cross-Attention establishes consistency and searches for correspondences between the source and target point clouds.}}
  \label{figurelabe8}
\end{figure}

\textbf{Geometric Self-attention}\quad To learn the global structural relationships within each point cloud, we apply a self-attention module. As illustrated in Figure~\ref{figurelabe8}(a), this module allows each superpoint to interact with all other superpoints within the same cloud, thereby capturing a comprehensive understanding of the global geometry.
\rev{Let the input feature matrix for the source point cloud be $\hat{F}^P \in \mathbb{R}^{N_P \times d}$. The self-attention mechanism updates these features to produce an output matrix $(\hat{F}^P)' \in \mathbb{R}^{N_P \times d}$.} Each output feature $(\hat{F}^P)'_i$ is a weighted sum of the projected input features:
\begin{equation}
(\hat{F}^P)'_i = \sum_{j=1}^{N_P} \alpha_{ij} (\hat{F}^P_j W_v) ,
\end{equation}
where $\alpha_{ij}$ are the attention weights derived from applying a softmax function to the attention scores $e_{ij}$.

The attention score $e_{ij}$ is computed based on the compatibility between a Query from superpoint $i$ and a Key from superpoint $j$. Crucially, our Key is a composite, incorporating both the learned feature $\hat{F}^P_j$ and the relative geometric embedding between the points, $E^P_{ij}$. This makes the attention mechanism spatially aware. The score is calculated as follows:
\begin{equation}
e_{ij} = \frac{ ( \hat{F}^P_i W_q ) (  \hat{F}^P_j W_k + E^P_{ij} W_g ^{\text{Key}} )^\top }{ \sqrt{d_k} } ,
\end{equation}
Here, $W_q, W_k, W_v,$ and $W_g$ are learnable weight matrices, and $d_k$ is the dimensionality of the key vectors.

To ensure effectiveness, the geometric embedding $E^P_{ij}$ is constructed from both distance and triplet angle information. 
We compute the angle embedding $E_{ij}^{A}$ using a sinusoidal function:
\begin{equation}
E^A_{ij} = \sin \left( \frac{\alpha_{ij}^x}{\sigma_\alpha} \right).
\end{equation}

Simultaneously, we compute the pairwise Euclidean distance $d_{ij} = \|p_i - p_j\|_2$. Similar to the angle embedding, the distance embedding $E_{ij}^{D}$ is obtained by encoding $d_{ij}$ via a sinusoidal function followed by a learnable Multi-Layer Perceptron.
Finally, the distance embedding $E_{ij}^{D}$ and angle embeddings are aggregated to form the final geometric embedding passed to the Key:
\begin{equation}
E^P_{ij} = E^D_{ij} W_D + \max_r \{ E^A_{ijr} W_A \} .
\end{equation}

By correlating feature-based similarities with their geometric arrangement, the self-attention mechanism produces a holistic and context-aware representation for each superpoint.

\textbf{Geometric Aggregation-attention}\quad
To effectively fuse the dense semantic cues from the 2D imagery into the sparse 3D geometric structures, we propose the Geometric Aggregation-Attention mechanism, illustrated in Figure~\ref{figurelabe8}(b). Unlike standard cross-attention which treats features as an unordered bag of tokens, our formulation explicitly models the spatial dependencies of both modalities to guide the feature retrieval process. \rev{In addition, $\mathcal{S}^P=\{p_i\}_{i=1}^{N_P}$ denote the set of 3D superpoint coordinates with associated geometric features $\hat{F}^P \in \mathbb{R}^{N_P \times d}$. Similarly, $\mathcal{G}^I = \{u_j\}_{j=1}^{M}$ represent the 2D pixel coordinates of the image patches, with visual features $\hat{F}^I \in \mathbb{R}^{M \times d}$. The attention mechanism functions as a retrieval process where 3D points (Queries) search for relevant visual context (Keys/Values) on the image plane. Crucially, to resolve ambiguities in repetitive textures or geometries, we inject spatial inductive biases directly into the attention formulation.} The attention score $e_{ij}$ between the $i$-th 3D superpoint and the $j$-th 2D image patch is computed as:
\begin{equation}
e_{ij} = \frac{ ( { \hat{F}^P_i W_q + E^P_{i} W_g } ) ( { \hat{F}^I_j W_k + E^I_{j} W_f } )^\top }{ \sqrt{d_k} } ,
\end{equation}
where $E_i^P \in \mathbb{R}^{d}$ and $E_j^I \in \mathbb{R}^{d}$ are the positional embeddings for the 3D superpoint coordinates and 2D pixel coordinates, respectively. \rev{Crucially, $W_f$ and $W_g$ project these spatial embeddings into a shared semantic space, enforcing geometric consistency between modalities. The final aggregated features $(\hat{F}^P)' \in \mathbb{R}^{N_P \times d}$ are obtained by a weighted sum of image values followed by a residual connection. $W_q$ and $W_k$ are linear projections for the content-based features of the point cloud and image, respectively.}

Finally, the attention weights are used to aggregate the image values, and the result is added back to the original point cloud features via a residual connection. This mechanism allows each 3D superpoint to selectively draw upon the most relevant cues from the 2D image, creating a fused feature representation that is far more discriminative than either modality alone.

\textbf{Geometric Cross-attention}\quad The cross-attention module is applied between the source and target point clouds to establish robust correspondences, as depicted in Figure~\ref{figurelabe8}(c). This mechanism allows each superpoint in the source cloud to attend to all superpoints in the target cloud, effectively searching for potential matches and modeling their geometric consistency.

\rev{In this interaction, the source point cloud ($\hat{F}^P$) provides the Query, while the target point cloud ($\hat{F}^Q$) provides the Key and Value. Let the features for the source and target superpoints be $\hat{F}^P \in \mathbb{R}^{N_P \times d}$ and $\hat{F}^Q \in \mathbb{R}^{N_Q \times d}$, respectively.} The attention score $e_{ij}$ between the $i$-th source superpoint and the $j$-th target superpoint is computed as:
\begin{equation}
e_{ij} = \frac{ ( { \hat{F}^P_i W_q + E^P_{i} W_g }) ({ \hat{F}^Q_j W_k + E^Q_{j} W_g } )^\top }{ \sqrt{d_k} } ,
\end{equation}
where $E^P_i$ and $E^Q_j$ are the respective geometric embeddings. The Value vectors are derived from the target features.

The resulting attention weights are then used to create a context vector by taking a weighted sum of the Values from the target cloud. This context vector is added to the source superpoint features via a residual connection, updating them with information about their most likely counterparts in the target cloud. This process is crucial for generating discriminative features that effectively capture the correspondences between the two point clouds for the final matching stage.

%\subsubsection{Super point Matching}
\textbf{Superpoint Matching}\quad After the hybrid attention mechanism, we have obtained the feature matrices $(\hat{F}^P)'$ and $(\hat{F}^Q)'$. The potential matching relationships between the two sets of points are measured using a similarity matrix $\mathit{S}$, where $S(m, n)$ represents the similarity between the $m$-th point in the source superpoint and the $n$-th point in the target superpoint.

If there are duplicate sampled point pairs, the similarity score for these pairs is set to $-\infty$ to avoid duplicate pairings from affecting the matching quality. Additionally, we add extra ``relaxation terms'' at the boundaries of the similarity matrix (initialized to $0$) to help the matrix converge.
Since certain regions of the point cloud have low geometric distinguishability, which can lead to ambiguous matches. We apply the Sinkhorn algorithm to perform dual normalization on the similarity matrix $\mathit{S}$. 

\rev{Given the similarity matrix $S \in \mathbb{R}^{N_P \times N_Q}$, matches for outliers (non-overlapping points) are handled by augmenting $S$ with a learnable dustbin. Specifically, we introduce a single learnable scalar parameter $z \in \mathbb{R}$, initialized to 0 and learned during training. This parameter acts as a flexible log-probability threshold for rejecting outliers. We construct the augmented score matrix $\bar{S} \in \mathbb{R}^{(N_P+1) \times (N_Q+1)}$ as:
\begin{equation}
\bar{S} = \begin{bmatrix} S & z \mathbf{e}_{N_P} \\ z \mathbf{e}_{N_Q}^\top & z \end{bmatrix},
\end{equation}
where $\mathbf{e}_{N}$ denotes a column vector of ones of size $N$. The last row and column correspond to the dustbin, absorbing probability mass from unmatched points. To transform $\bar{S}$ into a doubly stochastic probability matrix $Z$, we apply the Sinkhorn algorithm. We perform $L$ iterations (set to $L=50$ in our experiments) of alternating row and column normalization: $S_{row} = S_{ij} / \sum_k S_{ik}$ and $S_{col} = S_{ij} / \sum_k S_{kj}$. This iterative process ensures that the resulting matrix satisfies the constraints $\sum_j Z_{ij} = 1$ and $\sum_i Z_{ij} = 1$ (excluding the dustbin explicitly), providing a differentiable approximation to the linear assignment problem.}

\textbf{Select Corresponding Pairs}\quad From the normalized matrix, we select the $N_c$ largest elements as matching pairs, which form the final matching set $\tilde{\mathcal{C}}(l)$, where:
\begin{equation}
\tilde{\mathcal{C}}(l) = \left\{ \left( \hat{p}_{x_i}^l, \hat{p}_{y_j}^l \right) \mid (x_i, y_j) \in \text{top}_k \left( \hat{S} \right) \right\},
\end{equation}
and $\text{top}_k \left( \hat{S} \right)$ indicates the $k$ largest elements.

\subsection{Correspondence Module for Dense Point}
Through the superpoint matching module, we have obtained the correspondences between superpoints $\tilde{\mathcal{C}}(l)$. Building upon this coarse alignment, we then introduce a correspondence refinement module, which operates at a finer granularity to extract point-level correspondences.

For each selected pair of superpoints, we compute the similarity of their corresponding dense points and generate a relaxation matrix $\mathit{C}^i$.
We expand the original matrix $\mathit{C}^i$ into an enhanced matrix $\tilde{\mathit{C}}^i$. Then, we use the Sinkhorn algorithm to optimize the matrix, yielding a confidence matrix $\mathit{Z}_i$. We apply the top-k method to filter out the most reliable point pair correspondences.
Finally, we select the point pairs with the highest confidence from the confidence matrix:
\begin{equation}
\tilde{\mathcal{C}}_i = \left\{ \left( \hat{p}_i^r, \hat{q}_i^r \right) \mid (x_i, y_i) \in \text{top}_k \left( \hat{S} \right) \right\}.
\end{equation}

\subsection{Estimator Computation}

CMHANet provides correspondences with a high inlier ratio, allowing us to achieve high registration accuracy without relying on traditional estimators, while also reducing computational costs. Then the computation is divided into two stages:

\rev{In the local stage, we employ the Weighted Singular Value Decomposition (SVD) algorithm to compute the rigid transformation. We select SVD because it provides a differentiable, closed-form solution to the orthogonal Procrustes problem. This ensures the computationally efficient and optimal estimation of the rotation and translation that minimizes the weighted least-squares error for the given correspondences.} Consequently, we generate local transformations $(\mathit{R}_i, \mathit{t}_i)$ for each superpoint pair's matching result, which follows the relationship:
\begin{equation}
R_i, t_i = \min_{R, t} \sum_{(\tilde{p}_{x_j}, \tilde{q}_{y_j}) \in C_i} w_{j}^i \| R \cdot \tilde{p}_{x_j} + t - \tilde{q}_{y_j} \|_2^2 .
\end{equation}

\rev{In the global stage, we employ a 'Local-to-Global' verification strategy to ensure robustness without the nondifferentiability of RANSAC. From the set of generated local transformations $\{(R_i, t_i)\}$, we evaluate the quality of each candidate by counting its spatial inliers across the entire correspondence set $\mathcal{C}$. The selection criterion is defined as:
\begin{equation}
R^*, t^* = \max_{R_i, t_i} \sum_{(\tilde{p}_{x_j}, \tilde{q}_{y_j}) \in C} \left[ \left\| R_i \cdot \tilde{p}_{x_j} + t_i - \tilde{q}_{y_j} \right\|_2^2 < \tau_a \right],
\end{equation}
where $[\cdot]$ is the Iverson bracket~\cite{xu2024igreg}. The threshold $\tau_{a}$ represents the acceptable registration error margin (detailed in Sec 4.2).} This step effectively filters out incorrect local estimates caused by symmetric structures, ensuring that the final transformation is globally consistent.

\subsection{Overall Objective}
The training pipeline leverages a three-part loss function: (i) a coarse matching loss $\mathcal{L}_c$ for superpoint-level alignment, (ii) a fine matching loss $\mathcal{L}_f$ for point-level precision, and (iii) a cross-modal contrastive loss $\mathcal{L}_{\text{cmc}}$ to enforce modality-invariant feature representations.

%\subsubsection{Coarse Matching Loss}
\textbf{Coarse Matching Loss}\quad The coarse matching loss $\mathcal{L}_c$ optimizes the global correspondence of superpoints using an overlap-aware circle loss. We adopt a metric learning framework to address challenges such as inconsistent optimization for high-confidence positive pairs. Superpoint pairs with an overlap ratio exceeding 10\% are designated as positives, while pairs with no overlap are negatives. The anchor set $\mathcal{A}$ is constructed from source superpoints with at least one positive match in the target point cloud.

The overlap-aware circle loss for superpoint features is expressed as:

\begin{equation}
\begin{aligned}
    \mathcal{L}^P_c = \frac{1}{|\mathcal{A}|} \sum_{\hat{P}_i \in \mathcal{A}} \log \Bigg[ 1 + &\sum_{Q_j \in \epsilon_i} e^{\lambda^i_j \beta^p_{ij}(d^j_i - \Delta_p)} \cdot \\
    &\sum_{Q_k \in \epsilon_i}  e^{\beta^n_{i,k} (\Delta_n - d^k_i)} \Bigg]
\end{aligned},
\end{equation}

where $d^j_{i}$ is the feature distance, $\lambda^i_{j} = \sqrt{o^i_j}$ weights positive pairs based on overlap ratio $o^i_j$, and $\Delta p, \Delta n$ control the margin thresholds. By averaging $\mathcal{L}_c^P$ and its counterpart $\mathcal{L}_c^Q$ for the target point cloud, the final coarse matching loss becomes:$\mathcal{L}_c = \frac{\mathcal{L}_c^P + \mathcal{L}_c^Q}{2}$.
This loss encourages stronger alignment for highly overlapped superpoint pairs.

%\subsubsection{Fine Matching Loss}
\textbf{Fine Matching Loss}\quad The fine matching loss $\mathcal{L}_f$ minimizes the misalignment of point correspondences within each matched superpoint pair. For each ground-truth superpoint correspondence $\mathcal{C}^*$, the matched points $M_i$ are computed using a matching radius $\tau$. Remaining unmatched points are grouped into sets $I_i$ and $J_i$ for the source and target superpoints, respectively.

The point-level fine matching loss is defined as:
\begin{equation}
\begin{aligned}
    \mathcal{L}_{f,i} = - \sum_{(x, y) \in M_i} \log Z_i[x, y] 
              &- \sum_{x \in I} \log Z_i[x, m_i + 1] \\
              &- \sum_{y \in I} \log Z_i[n_i + 1, y],
\end{aligned}
\end{equation}
where $Z_i[x,y]$ is the predicted probability of the $x$-th source point matching the $y$-th target point. The complete fine matching loss is given by:
\begin{equation}
\mathcal{L}_f = \frac{1}{N} \sum_{i=1}^g \mathcal{L}_{f,i}.
\end{equation}

This loss refines local geometric alignment, ensuring accurate point-wise correspondence.

%\subsubsection{Cross-Modal Contrastive Loss}
\textbf{Cross-Modal Contrastive Loss}\quad To ensure consistent feature representations across modalities, we introduce a cross-modal contrastive loss $\mathcal{L}_{\text{cmc}}$. This loss enforces that features of source and target modalities are pulled closer in the embedding space.

The process operates on the extracted features of the input point cloud and image pair. \rev{Let $N_P$ denote the number of sampled superpoints in the current scene (as defined in Section 3.2). We construct the contrastive loss at the superpoint level rather than the scene level. Specifically, we extract the geometric features $F^P \in \mathbb{R}^{N_P \times d}$ and the corresponding image features $F^I \in \mathbb{R}^{N_P \times d}$ for the $N_P$ superpoints. A pairwise similarity matrix $S$ of size $N_P \times N_P$ is computed, where the element $s[i,j]$ represents the similarity score between the geometric feature of the $i$-th superpoint and the image feature of the $j$-th superpoint.The cross-modal contrastive loss is defined as:
\begin{equation}
\mathcal{L}_{\text{cmc}} = - \frac{1}{N_P} \sum_{i=1}^{N_P} \log \frac{\exp(s[i,i])}{\sum_{j=1}^{N_P} \exp(s[i,j])}.
\end{equation}}

\rev{In this formulation, the diagonal elements represent positive pairs, while off-diagonal elements serve as negative pairs. This ensures the loss remains effective even when the training batch size is set to 1.}
This loss emphasizes instance-level coherence across modalities, encouraging features from corresponding instances in source and target point clouds to remain consistent.

%\subsubsection{Overall Loss Function}
\textbf{Overall Loss Function}\quad The final loss function integrates the above components to jointly optimize global alignment, local precision, and cross-modal feature consistency:
\begin{equation}
\mathcal{L} = \mathcal{L}_c + \mathcal{L}_f + \lambda \mathcal{L}_{\text{cmc}},
\end{equation}
where $\lambda$ is a hyperparameter.
These loss functions orchestrate a coarse-to-fine optimization strategy, enabling the model to robustly align source and target point clouds.

\begin{table}[t]
\centering
\small
\caption{Evaluation results of our method on 3DMatch and 3DLoMatch datasets. The results are reported for two key metrics: Feature Matching Recall (\%) and Registration Recall (\%). The evaluation is conducted across varying sample sizes (5000, 2500, 1000, 500 and 250) to assess the robustness of each method under different constraints.}
%CMHANet consistently achieves top performance, with the highest Feature Matching Recall and Registration Recall.
\label{table1}
\renewcommand{\arraystretch}{0.70}
\setlength{\tabcolsep}{4pt} % 缩小列间距
\resizebox{0.8\textwidth}{!}{ % 自动缩放到单栏宽度
\begin{tabular}{l|c|c|c|c|c|c|c|c|c|c}
\toprule
 & \multicolumn{5}{c|}{3DMatch} & \multicolumn{5}{c}{3DLoMatch} \\
\# Samples & 5000 & 2500 & 1000 & 500 & 250 & 5000 & 2500 & 1000 & 500 & 250 \\
\midrule
\multicolumn{11}{c}{Feature Matching Recall (\%) $\uparrow$} \\
\midrule
PerfectMatch\rev{~\citep{gojcic2019perfect}} & 95.0 & 94.3 & 92.9 & 90.1 & 82.9 & 63.6 & 61.7 & 53.6 & 45.2 & 34.2 \\
FCGF\rev{~\citep{choy2019fully}} & 97.4 & 97.3 & 97.0 & 96.7 & 96.6 & 76.6 & 75.4 & 74.2 & 71.7 & 67.3 \\
D3Feat\rev{~\citep{bai2020d3feat}} & 95.6 & 94.3 & 95.6 & 94.3 & 91.0 & 66.7 & 64.7 & 61.6 & 60.0 & 56.6 \\
SpinNet\rev{~\citep{ao2021spinnet}} & 97.6 & 97.2 & 96.8 & 95.5 & 96.5 & 75.3 & 74.9 & 72.5 & 70.0 & 63.6 \\
Predator\rev{~\citep{huang2021predator}} & 96.6 & 96.6 & 96.5 & 96.3 & 95.3 & 78.6 & 76.7 & 75.7 & 75.7 & 72.9 \\
YOHO\rev{~\citep{wang2022you}} & 98.2 & 97.6 & 97.5 & 97.7 & 96.0 & 79.4 & 78.1 & 76.3 & 73.8 & 69.1 \\
CoFiNet\rev{~\citep{yu2021cofinet}} & 98.1 & 98.3 & 98.1 & 98.2 & 98.3 & 83.1 & 83.5 & 83.3 & 83.1 & 82.6 \\
%GeoTransformer & 97.9 & 97.9 & 97.9 & 97.9 & 97.6 & \textbf{88.3} & \textbf{88.6} & \textbf{88.8} & \textbf{88.6} & \textbf{88.3} \\
% IMFNet & \underline{98.5} & \underline{98.5} & \underline{98.2} & 98.1 & 97.5 & - & - & - & - & - \\
% PCR-CG & 97.4 & 97.5 & 97.7 & 97.3 & 97.6 & 80.4 & 82.2 & 82.6 & 83.2 & 82.8 \\
RoReg\rev{~\citep{wang2023roreg}} & 98.2 & 97.9 & 98.2 & 97.8 & 97.2 & 82.1 & 82.1 & 81.7 & 81.6 & 80.2 \\
OIF-PCR\rev{~\citep{yang2022one}} & 98.1 & 98.1 & 97.9 & 98.4 & \rev{\textbf{98.4}} & 84.6 & 85.2 & 85.5 & 86.6 & 87.0 \\
CMHANet (\textit{ours}) & \textbf{98.6} & \textbf{98.6} & \textbf{98.6} & \textbf{98.6} & \textbf{98.4} & \textbf{87.7} & \textbf{87.5} & \textbf{87.5} & \textbf{87.5} & \textbf{87.4} \\
\midrule
\multicolumn{11}{c}{Inlier Ratio (\%) $\uparrow$} \\
\midrule
PerfectMatch\rev{~\citep{gojcic2019perfect}} & 36.0 & 32.5 & 26.4 & 21.5 & 16.4 & 11.4 & 10.1 & 8.0 & 6.4 & 4.8 \\
FCGF\rev{~\citep{choy2019fully}} & 56.8 & 54.1 & 48.7 & 42.5 & 34.1 & 21.4 & 20.0 & 17.2 & 14.8 & 11.6 \\
D3Feat\rev{~\citep{bai2020d3feat}} & 39.0 & 38.8 & 40.4 & 41.5 & 41.8 & 13.2 & 13.1 & 14.0 & 14.6 & 15.0 \\
SpinNet\rev{~\citep{ao2021spinnet}} & 47.5 & 44.7 & 39.4 & 33.9 & 27.6 & 20.5 & 19.0 & 16.3 & 13.8 & 11.1 \\
Predator\rev{~\citep{huang2021predator}} & 58.0 & 58.4 & 57.1 & 54.1 & 49.3 & 26.7 & 28.1 & 28.3 & 27.5 & 25.8 \\
YOHO\rev{~\citep{wang2022you}} & 64.4 & 60.7 & 55.7 & 46.4 & 41.2 & 25.9 & 23.3 & 22.6 & 18.2 & 15.0 \\
CoFiNet\rev{~\citep{yu2021cofinet}} & 49.8 & 51.2 & 51.9 & 52.2 & 52.2 & 24.4 & 25.9 & 26.7 & 26.8 & 26.9 \\
OT-CA\rev{~\citep{guo2024learning}} & 59.5 & 61.5 & 60.9 & 61.7 & 61.7 & 30.5 & 30.6 & 30.5 & 34.9 & 30.2 \\
OIF-PCR\rev{~\citep{yang2022one}} & 62.3 & 65.2 & 66.8 & 67.1 & 67.5 & 27.5 & 30.0 & 31.2 & 32.6 & 33.1 \\
%GeoTransformer & \textbf{71.9} & \underline{75.2} & \underline{76.0} & \underline{82.2} & \underline{85.1} & \underline{43.5} & \underline{45.3} & \underline{46.2} & \underline{52.9} & \underline{57.7} \\
CMHANet (\textit{ours}) & \textbf{71.4} & \textbf{77.7} & \textbf{78.9} & \textbf{83.5} & \textbf{86.2} & \textbf{43.7} & \textbf{47.4} & \textbf{52.3} & \textbf{55.6} & \textbf{58.3} \\
\midrule
\multicolumn{11}{c}{Registration Recall (\%) $\uparrow$} \\
\midrule
PerfectMatch\rev{~\citep{gojcic2019perfect}} & 78.4 & 76.2 & 71.4 & 67.6 & 50.8 & 33.0 & 29.0 & 23.7 & 11.0 & 15.9 \\
FCGF\rev{~\citep{choy2019fully}} & 85.1 & 83.4 & 82.7 & 81.6 & 76.6 & 74.4 & 70.4 & 67.2 & 63.4 & 61.4 \\
D3Feat\rev{~\citep{bai2020d3feat}} & 81.6 & 84.5 & 83.4 & 82.7 & 77.9 & 69.4 & 60.2 & 58.5 & 51.6 & 42.6 \\
SpinNet\rev{~\citep{ao2021spinnet}} & 86.6 & 85.8 & 85.5 & 83.5 & 79.7 & 68.9 & 59.6 & 56.8 & 50.0 & 49.1 \\
Predator\rev{~\citep{huang2021predator}} & 89.0 & 89.9 & 90.6 & 88.5 & 86.6 & 61.2 & 61.2 & 62.4 & 60.8 & 58.1 \\
YOHO\rev{~\citep{wang2022you}} & 90.8 & 90.3 & 89.1 & 88.6 & 84.5 & 67.5 & 68.0 & 62.2 & 63.4 & 61.9 \\
CoFiNet\rev{~\citep{yu2021cofinet}} & 89.3 & 88.9 & 88.4 & 87.4 & 87.0 & 67.5 & 66.2 & 64.2 & 63.1 & 61.0 \\
%GeoTransformer & \underline{92.0} & \underline{91.8} & \underline{91.8} & \underline{91.4} & \underline{91.2} & \underline{75.0} & \underline{74.8} & \underline{74.2} & \underline{74.1} & \underline{73.5} \\
% IMFNet & 91.0 & - & - & - & - & - & - & - & - & - \\
% PCR-CG & 89.4 & 90.7 & 90.0 & 88.7 & 86.8 & 66.3 & 67.2 & 69.0 & 68.5 & 65.0 \\
FeatSync\rev{~\citep{hu2024featsync}} & 90.2 & 90.3 & 90.6 & 91.2 & 91.1 & - & - & - & - & - \\

CMHANet (\textit{ours}) & \textbf{92.4} & \textbf{92.4} & \textbf{91.9} & \textbf{92.0} & \textbf{91.7} & \textbf{75.5} & \textbf{75.1} & \textbf{74.8} & \textbf{74.6} & \textbf{74.6} \\
\bottomrule
\end{tabular}
} % end resizebox
\end{table}

\section{EXPERIMENT}

In this section, we conduct relevant experiments to validate the effectiveness of our model.

\begin{table}[t]
\centering
\caption{RRE and RTE on 3DMatch (3DM) and 3DLoMatch (3DLM) benchmarks. We report Relative Rotation Error (RRE, °) and Relative Translation Error (RTE, m) for various methods using RANSAC-50k or RANSAC-free estimators. CMHANet achieves the lowest RRE and RTE on both datasets.}
\label{table3} 
\renewcommand{\arraystretch}{0.8}
\setlength{\tabcolsep}{4pt} % 缩小列间距
\resizebox{\linewidth}{!}{ % 自动缩放到单栏宽度
\begin{tabular}{l|c|cc|cc}
\toprule
\multirow{2}{*}{Model} & \multirow{2}{*}{Estimator} & \multicolumn{2}{c|}{3DMatch} & \multicolumn{2}{c}{3DLoMatch} \\
 & & RRE (°) & RTE (m) & RRE (°) & RTE (m) \\
\midrule
Predator \citep{huang2021predator} & RANSAC-\textit{50k} & 2.029 & 0.064 & 3.048 & 0.093 \\
CoFiNet \citep{yu2021cofinet} & RANSAC-\textit{50k} & 2.002 & 0.064 & 3.271 & 0.090 \\
PCR-CG \citep{zhang2022pcr} & RANSAC-\textit{50k} & 1.993 & 0.061 & 3.002 & 0.087 \\
\rev{MAC~\citep{zhang20233d}} & RANSAC-free & 2.003 & 0.065 & 3.010 & 0.085 \\
\rev{Hybrid~\citep{jiang2025hybrid}} & RANSAC-free & 1.992 & 0.061 & 3.002 & 0.087 \\
GeoTransformer \citep{qin2022geometric} & RANSAC-free & 1.772 & 0.061 & 3.010 & 0.087 \\
\rev{BUFFER \citep{ao2023buffer}} & RANSAC-free & 1.850 & 0.061 & 3.090 & 0.101 \\
\rev{Semreg \citep{fung2024semreg}} & RANSAC-free & 1.830 & 0.061 & \textbf{2.782} & 0.095 \\
CMHANet (\textit{ours}) & RANSAC-free & \textbf{1.764} & \textbf{0.060} & 2.839 & \textbf{0.084} \\
\bottomrule
\end{tabular}
} % end resizebox
\end{table}

\begin{table}[t]
\centering
\caption{Comparison With Multimodal Point Cloud Registration}
\label{tab:multimodal}
\begin{tabular}{l|cc|cc}
\hline
\multirow{2}{*}{Model} & \multicolumn{2}{c|}{3DMatch} & \multicolumn{2}{c}{3DLoMatch} \\
 & FMR & RR & FMR & RR \\
\hline
IMFNet\citep{huang2022imfnet} & 98.5 & 91.0 & 80.5 & 48.4 \\
PCR-CG\citep{zhang2022pcr} & 97.4 & 89.4 & 80.4 & 66.3 \\
CMHANet (\textit{ours}) & \textbf{98.6} & \textbf{92.4} & \textbf{87.7} & \textbf{75.5} \\
\hline
\end{tabular}
\end{table}

\begin{table}[t]
\centering
\caption{Comparison of Model Registration Performance with Different Estimators and Sample sizes On 3DMatch(3DM) and 3DLomatch(3DLM). We report Registration Recall (RR, \%) and computation times (Model, Pose, and Total, in seconds) for various methods. Evaluations are conducted with RANSAC-50k, Weighted SVD, and LGR to assess robustness and efficiency. CMHANet consistently achieves superior performance in both metrics.}
\label{table2}

\renewcommand{\arraystretch}{0.8} % 行距稍微压缩
\setlength{\tabcolsep}{4pt} % 缩小列间距

\resizebox{\linewidth}{!}{ % 自动缩放到单栏
\begin{tabular}{l|c|c|cc|ccc}
\toprule
\multirow{2}{*}{Model} & \multirow{2}{*}{Estimator} & \multirow{2}{*}{\# Samples} & \multicolumn{2}{c|}{RR (\%)} & \multicolumn{3}{c}{Times (s)} \\
 & & & 3DM & 3DLM & Model & Pose & Total \\
\midrule
FCGF \citep{choy2019fully} & RANSAC-50k & 5000 & 85.1 & 40.1 & 0.052 & 3.326 & 3.378 \\
D3Feat \citep{bai2020d3feat} & RANSAC-50k & 5000 & 81.6 & 37.2 & 0.024 & 3.088 & 3.112 \\
SpinNet \citep{ao2021spinnet} & RANSAC-50k & 5000 & 88.6 & 59.8 & 0.248 & 0.388 & 60.636 \\
Predator \citep{huang2021predator} & RANSAC-50k & 5000 & 89.0 & 59.8 & 0.032 & 5.120 & 5.152 \\
CoFiNet \citep{yu2021cofinet} & RANSAC-50k & 5000 & 89.3 & 67.5 & 0.115 & 1.807 & 1.922 \\
%GeoTransformer \citep{qin2022geometric} & RANSAC-50k & 5000 & \underline{92.0} & \underline{75.0} & 0.075 & 1.558 & \textbf{1.633} \\
CMHANet (\textit{ours}) & RANSAC-50k & 5000 & \textbf{92.4} & \textbf{75.5} & 0.144 & 1.732 & 1.876 \\
\midrule
FCGF \citep{choy2019fully} & Weighted SVD & 250 & 42.1 & 3.9 & 0.052 & 0.008 & 0.056 \\
D3Feat \citep{bai2020d3feat} & Weighted SVD & 250 & 37.4 & 2.8 & 0.024 & 0.004 & \textbf{0.032} \\
SpinNet \citep{ao2021spinnet} & Weighted SVD & 250 & 38.0 & 5.0 & 0.248 & 0.009 & 0.254 \\
Predator \citep{huang2021predator} & Weighted SVD & 250 & 50.0 & 4.6 & 0.032 & 0.009 & 0.041 \\
CoFiNet \citep{yu2021cofinet} & Weighted SVD & 250 & 63.0 & 21.6 & 0.115 & 0.003 & 0.118 \\
%GeoTransformer \citep{qin2022geometric} & Weighted SVD & 250 & \underline{86.5} & \underline{59.9} & 0.075 & 0.010 & 0.078 \\
CMHANet (\textit{ours}) & Weighted SVD & 250 & \textbf{87.9} & \textbf{60.4} & 0.144 & 0.004 & 0.148 \\
\midrule
CoFiNet \citep{yu2021cofinet} & LGR & all & 87.6 & 64.8 & 0.115 & 0.023 & 0.188 \\
%GeoTransformer \citep{qin2022geometric} & LGR & all & 91.5 & 74.0 & 0.075 & 0.013 & \textbf{0.088} \\
CMHANet (\textit{ours}) & LGR & all & \textbf{91.9} & \textbf{74.2} & 0.144 & 0.042 & 0.186 \\
\bottomrule
\end{tabular}
} % end resizebox
\end{table}

\subsection{\rev{Benchmarks and Metrics}}
%\subsubsection{Dataset}
\textbf{Dataset}\quad Both 3DMatch\citep{zeng20173dmatch} and 3DLoMatch\citep{huang2021predator} are built using RGB-D data, with their point clouds generated by processing 50 consecutive frames of RGB and depth images. To construct the multimodal dataset, we extract the corresponding images for each point cloud and pair them with the point clouds to ensure both describe the same scene.
3DMatch contains 62 scenes, with 46 used for training, 8 for validation, and 8 for testing. For 3DMatch, any registered point cloud pair has an overlap rate of over 30\%, whereas for 3DLoMatch, the overlap rate is between 10\% and 30\%. \rev{To evaluate the cross-domain generalization capability of our method, we additionally employ the TUM RGB-D SLAM dataset\citep{sturm2012benchmark}. This dataset distincts from the 3DMatch domain. Following standard protocols, we sample point cloud pairs at fixed intervals (every 5 frames for Camera 1 and every 20 frames for Camera 2) to generate testing pairs with varying overlap.}

%\subsubsection{Metrics and Main Results}
\textbf{Metrics and Estimators}\quad 
\rev{To comprehensively evaluate the model, we employ Feature Matching Recall (FMR), Registration Recall (RR), Relative Rotation Error (RRE), and Relative Translation Error (RTE) as evaluation metrics. For the pose estimation stage, Random Sample Consensus (RANSAC), weighted SVD, and Local-to-Global Registration (LGR) are used as estimators to compute the rigid transformation.}

\subsection{\rev{Experiment Details}}

The implementation details are based on PyTorch and trained on an NVIDIA RTX 3090 GPU. The model is trained for 50 epochs on 3DMatch using the Adam optimizer. The batch size is set to 1, and the weight decay rate is set to $10^{-6}$. The learning rate starts from $10^{-4}$ and decays exponentially by a factor of 0.05 every epoch on 3DMatch. Additionally, the matching radius threshold on 3DMatch is set $\tau_a = 5 \, \text{cm}$ and $\lambda=0.5$.

\textbf{Baseline Implementation}\quad To ensure a fair and rigorous comparison, we evaluated all sota baseline methods \citep{huang2021predator,yu2021cofinet,guo2024learning,hu2024featsync} using their official open-source implementations and pre-trained models provided by the original authors. \rev{We adhered to the standard parameter settings recommended in their respective papers for the 3DMatch/3DLoMatch benchmarks. Furthermore, for the runtime analysis presented in Table ~\ref{table2}, all baseline models were executed on the same hardware environment (NVIDIA RTX 3090 GPU) to accurately measure inference latency under identical conditions.}

\subsection{Main Results}
\textbf{Quantitative Analysis on Standard Benchmarks}\quad
We first evaluate the registration performance on the 3DMatch and 3DLoMatch benchmarks. Table~\ref{table1} presents the FMR and RR. CMHANet achieves state-of-the-art performance, with a Registration Recall of 92.4\% on 3DMatch and 75.5\% on the challenging 3DLoMatch dataset. This demonstrates the effectiveness of our method in identifying robust correspondences even in low-overlap scenarios.

\begin{figure*}[t]
  \centering

  \includegraphics[width=1\linewidth]{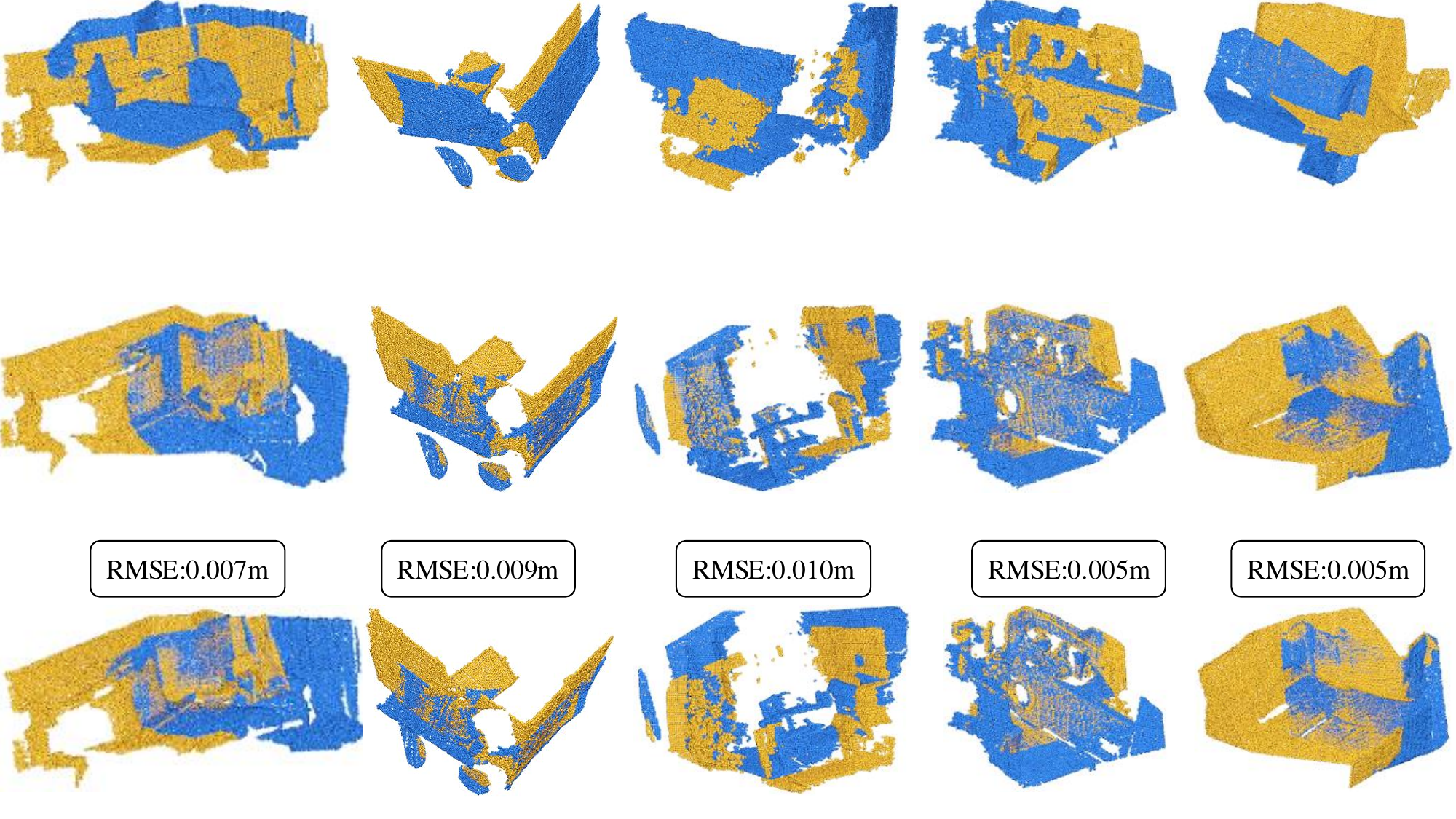}
  \caption{The result of CMHANet. \rev{We visualized representative scenes from the 3DMatch dataset.} The first row shows the initial arrangement of the two point clouds in their raw, unaligned positions. The middle row depicts the ground truth alignment of the two point clouds, representing the ideal registration outcome. The bottom row showcases the registration results predicted by CMHANet, offering a direct comparison to both the initial misalignment and the ground truth. We also report the Root Mean Square Error (RMSE) values for each scene, quantifying the spatial deviation between CMHANet’s predicted alignments and the ground truth.}
  \label{figurelabe4}
\end{figure*}

For FMR, CMHANet achieves a significant improvement on both benchmarks. On 3DMatch, CMHANet outperforms existing methods, which demonstrates its ability to identify robust correspondences. On 3DLoMatch, the reliance on multimodal image features becomes a limitation, as the low-overlap nature of the dataset reduces the contribution of visual information, leading to a slight drop. Figure.~\ref{figurelabe4} provides a series of the registration results of our model.

\rev{Complementing the recall metrics, Table~\ref{table3} reports the registration accuracy in terms of RRE and RTE.} CMHANet achieves the lowest errors on both datasets (RRE: 1.764°, RTE: 0.060m on 3DMatch), confirming that our cross-modal fusion not only improves the success rate but also the precision of the alignment.

\textbf{Comparison with Multimodal Methods}\quad
We also compare CMHANet with recent multimodal point cloud registration methods, PCR-CG \citep{zhang2022pcr} and IMFNet \citep{huang2022imfnet}, as shown in Table~\ref{tab:multimodal}. Our method significantly outperforms these baselines (e.g., +9.2\% RR on 3DLoMatch vs. PCR-CG), validating the superiority of our Hybrid Attention mechanism in fusing 2D and 3D features.

\textbf{Robustness and Efficiency Analysis}\quad
We further evaluate the robustness of CMHANet under different pose estimators (RANSAC, Weighted SVD, LGR). The results shown in Table~\ref{table2}, demonstrate that CMHANet outperforms baseline methods across all estimators in RR on both 3DMatch and 3DLoMatch datasets.

Notably, CMHANet achieves 91.9\% RR on 3DMatch and 74.2\% RR on 3DLoMatch when using LGR. Furthermore, its LGR-based results achieve competitive accuracy compared to RANSAC while being over 100 times faster in pose estimation.
We also analyze processing time, focusing on feature extraction, pose estimation time. While CMHANet requires slightly longer feature extraction due to image feature encoding and the integration of multimodal features. The improved quality of correspondences contributes to faster convergence during pose computation, ensuring the total runtime remains competitive.
% These results emphasize the advantages of CMHANet in terms of both accuracy and speed. By leveraging its hybrid attention mechanism and multimodal feature fusion, CMHANet achieves robust and efficient registration.

\textbf{The Generalization of CMHANet}\quad
\begin{table*}[t]
\centering
\caption{\rev{Registration Results on Eight Sequences from the TUM RGB-D SLAM Dataset. We report the Average RMSE ($\times 10^{-2}$). Our method (CMHANet) is trained on 3DMatch and directly tested on TUM without fine-tuning.}}
\label{table:generalization_tum}
\resizebox{\textwidth}{!}{
\begin{tabular}{l|cccccccc|c}
\toprule
Method & xyz & 360 & teddy & desk & plant & dishes & coke & flower & mean \\
\midrule
ICP~\cite{besl1992method} & 2.1 & 5.1 & 2.1 & 2.3 & 1.6 & 3.7 & 3.1 & 2.7 & 2.8 \\
AA-ICP~\cite{pavlov2018aa} & 2.1 & 5.1 & 2.1 & 2.3 & 1.6 & 3.7 & 3.1 & 2.7 & 2.8 \\
Sparse ICP~\cite{bouaziz2013sparse} & 1.6 & 4.8 & 1.8 & 1.8 & 0.88 & 3.9 & 3.3 & 3.2 & 2.66 \\
Robust ICP~\cite{zhang2021fast} & 0.5 & 2.2 & 1 & 1.2 & 0.65 & 3.2 & 2.4 & 2.4 & 1.69 \\
Teaser++~\cite{yang2020teaser} & 3.4 & 20 & 11 & 8 & 6.1 & 21 & 23 & 20 & 14.06 \\
DCP~\cite{wang2019deep} & 6.5 & 10 & 6.6 & 7 & 5.6 & 7.4 & 7.5 & 6.3 & 7.11 \\
DGR~\cite{choy2020deep} & 0.6 & 1.4 & 1 & 1.2 & 0.71 & 2.6 & 2.1 & 1.9 & 1.44 \\
\midrule
CMHANet (Ours) & \textbf{0.2} & \textbf{0.3} & \textbf{0.51} & \textbf{0.6} & \textbf{0.32} & \textbf{1.8} & \textbf{1.4} & \textbf{1.0} & \textbf{0.76} \\
\bottomrule
\end{tabular}
}
\end{table*}
\rev{To address the concern regarding cross-domain generalization and to demonstrate that our model has not overfitted to the 3DMatch domain, we conducted a zero-shot generalization experiment on the TUM RGB-D SLAM dataset~\cite{sturm2012benchmark}.
We utilized the model trained solely on the 3DMatch dataset and directly evaluated it on the TUM dataset without any fine-tuning. We selected 8 representative sequences captured by different cameras. For sequences xyz, 360, teddy, desk, and plant (Camera 1), we used one in every 5 frames. For dishes, coke, and flower bouquet (Camera 2), we used one in every 20 frames. Following standard practices, the evaluation metric is the Root Mean Square Error (RMSE) of the estimated transformation.}

\rev{The quantitative results are summarized in Table~\ref{table:generalization_tum}. As observed, our method demonstrates generalization capabilities on this unseen dataset. Specifically, our method achieves a mean RMSE of 0.76 ($10^{-2}$). Furthermore, CMHANet surpasses robust optimization-based methods such as Robust ICP (1.69) and Teaser++ (14.06). Notably, on high-quality sequences like xyz and 360, our method achieves extremely low errors of 0.2 and 0.3, respectively. Even on the more challenging Camera 2 sequences (dishes, coke), CMHANet maintains high stability.}

\begin{table}[t]
\centering
\scriptsize % Adjust the font size for the table
\caption{\rev{Ablation experiments of the CMHANet. The results are measured in \%. We also assess the impact of including ($\checkmark$) or excluding (-) our custom Loss function, Hybrid Attention (HA) module, and Image Module (IM) on PIR, FMR, IR, and RR metrics. Best performance is highlighted in bold.}}
\label{table4}
\resizebox{\columnwidth}{!}{ % Resize the table to fit the single column width
\begin{tabular}{ccc|c|c|c|c|c|c|c|c}
\toprule
\multicolumn{3}{c}{Module} & \multicolumn{4}{c}{3DMatch} & \multicolumn{4}{c}{3DLoMatch} \\ Loss & HA & IM & PIR & FMR & IR & RR & PIR & FMR & IR & RR \\
\midrule
$\checkmark$ & - & - & 83.8 & 97.4 & 68.2 & 89.9 & 50.9 & 86.0 & 41.2 & 71.9 \\
- & - & $\checkmark$ & 84.2 & 97.8 & 65.9 & 90.1 & 51.2 & 86.3 & 40.8 & 72.1 \\
- & $\checkmark$ & $\checkmark$ & 85.1 & 98.5 & 69.2 & 91.4 & 51.7 & 86.8 & 42.9 & 74.2 \\
$\checkmark$ & - & $\checkmark$ & 84.0 & 98.2 & 68.8 & 90.5 & 52.2 & 86.5 & 41.8 & 72.4 \\
$\checkmark$ & $\checkmark$ & $\checkmark$ & \textbf{86.8} & \textbf{98.6} & \textbf{70.7} & \textbf{92.4} & \textbf{54.8} & \textbf{87.7} & \textbf{43.1} & \textbf{75.5} \\
\bottomrule
\end{tabular}
}
\end{table}

\textbf{\rev{Qualitative Results}}\quad
Finally, Figure~\ref{figurelabe4} visualizes representative registration results. The qualitative comparisons show that CMHANet successfully aligns point clouds in complex indoor scenes where geometric features alone might be ambiguous.

\rev{While CMHANet demonstrates robust performance across most scenarios, we acknowledge certain limitations in extreme conditions. Specifically, in cases characterized by extremely low overlap ratios (<10\%) or strictly featureless, flat surfaces where both geometric and textural cues are scarce, the model may occasionally exhibit registration deviations. Addressing robustness in these extreme edge cases remains a focus for our future work.}

\subsection{Ablation Experiment}
%\textbf{Ablation Experiment}\quad 

\rev{In this section, we verify the effectiveness of the key components in CMHANet. We analyze the contributions of the custom Loss function, the Hybrid Attention (HA) module, and the Image Module (IM).}

\textbf{\rev{Impact of Key Components}}\quad
We first assess the individual contribution of each major component by removing them from the full model. \rev{The overall results are summarized in Table~\ref{table4}.}
These evaluations were performed on the 3DMatch and 3DLoMatch datasets, employing metrics such as Patch Inlier Ratio (PIR), FMR, IR, and RR, all reported in percentages. The complete CMHANet model, integrating all components (row 5), establishes the performance benchmark.

The efficacy of the Loss is evident when comparing the full model to its configuration without this (row 3 vs. row 5). Its omission led to a notable performance decrease, highlighting the loss function's role in refining correspondence quality.
Similarly, the critical contribution of the HA module is demonstrated by its exclusion (row 4 vs. row 5). This resulted in significant performance degradation across both datasets.
The IM Module also demonstrates a positive impact on performance. Comparing configurations with and without the IM ($e.g.$, row 4 vs. row 1, where Loss is present but HA is not) shows consistent metric improvements upon IM's inclusion.

\begin{table}[t]
\centering
\caption{\rev{Ablation Experiments on Image Module in CMHANet. Best performance is highlighted in bold. The second-best performance results are indicated by underlines.}}
\label{tab:ablation}
\begin{tabular}{l|cccc|cccc}
\hline
\multirow{2}{*}{Model} & \multicolumn{4}{c|}{3DMatch} & \multicolumn{4}{c}{3DLoMatch} \\
 & PIR & FMR & IR & RR & PIR & FMR & IR & RR \\
\hline
w/o IM Module & 83.8 & 97.4 & 68.2 & 89.9 & 50.9 & 86.0 & 41.2 & 71.9 \\
Resnet-34  & 84.9 & 97.8 & 68.7 & 90.9 & 51.2 & 86.3 & 41.8 & 73.1 \\
Resnet-101  & \textbf{87.0} & \underline{98.5} & \textbf{71.2} & \underline{92.4} & \underline{54.7} & \underline{87.2} & \underline{42.9} & \underline{75.2}\\
CMHANet & \underline{86.8} & \textbf{98.6} & \underline{70.7} & \textbf{92.4} & \textbf{54.8} & \textbf{87.7} & \textbf{43.1} & \textbf{75.5} \\
\hline
\end{tabular}
\end{table}

\textbf{\rev{Analysis of Image Module Architecture}}\quad 
To further investigate the Image Module, we conduct a series of ablation experiments to validate the effectiveness of our proposed cross-modal fusion strategy and to analyze the impact of the IM Module architecture within CMHANet. Our full model utilizes a ResUNet-50 backbone for image feature extraction. \rev{The detailed results of these experiments are presented in Table \ref{tab:ablation}.} We first investigate the effectiveness of our core cross-modal design by removing the IM Module entirely. In this configuration, denoted as w/o IM Module, CMHANet is modified to be a single-modal method that relies solely on geometric information for registration. As shown in Table \ref{tab:ablation}, removing the image features leads to a significant performance degradation across all metrics.

Next, we analyze the impact of the image encoder's depth on registration performance. We replaced our standard ResUNet-50 backbone with both a lighter ResNet-34 and a deeper ResNet-101. The results indicate a clear trend: the representational power of the image encoder directly influences the final registration quality. Using the lighter ResNet-34 leads to a noticeable drop in performance compared to our full model. Conversely, the deeper ResNet-101 backbone yields results that are highly competitive with our ResUNet-50 configuration. However, our chosen CMHANet with ResUNet-50 achieves the highest or second-highest scores across all metrics, and notably secures the best RR on both datasets. This suggests that ResUNet-50 strikes an effective balance between powerful feature extraction and model efficiency for this task.

\begin{table}[t]
\centering
\caption{\rev{Ablation Experiments on Hybrid Attention Module in CMHANet. Best performance is highlighted in bold.}}
\label{tab:ablation_aggr}
\begin{tabular}{l|cccc|cccc}
\hline
\multirow{2}{*}{Model} & \multicolumn{4}{c|}{3DMatch} & \multicolumn{4}{c}{3DLoMatch} \\
 & PIR & FMR & IR & RR & PIR & FMR & IR & RR \\
\hline
w/o HA Module & 84.0 & 98.2 & 68.8 & 90.5 & 52.2 & 86.5 & 41.8 & 72.4 \\
w/o Aggre Att & 85.9 & 98.3 & 70.2 & 91.3 & 53.7 & 86.7 & 42.8 & 73.6 \\
CMHANet & \textbf{86.8} & \textbf{98.6} & \textbf{70.7} & \textbf{92.4} & \textbf{54.8} & \textbf{87.7} & \textbf{43.1} & \textbf{75.5} \\
\hline
\end{tabular}
\end{table}

\textbf{\rev{Analysis of Hybrid Attention Internal Design}}\quad 
Finally, to verify the effectiveness of our proposed HA module, we conducted a detailed ablation study. The HA module is composed of iteratively applied self-attention, aggregation-attention and cross-attention mechanisms. We analyze the contribution of the module as a whole, as well as the specific importance of its cross-modal fusion component. \rev{The results are summarized in Table \ref{tab:ablation_aggr}.}

We first evaluate the contribution of the entire HA module by removing it completely (w/o HA Module) and presumably replacing it with a more direct feature comparison method. As shown in the first row of Table \ref{tab:ablation_aggr}, this modification leads to a substantial performance drop across both datasets. Notably, the primary RR metric decreases by 1.9\% on 3DMatch (from 92.4\% to 90.5\%) and a significant 3.1\% on 3DLoMatch (from 75.5\% to 72.4\%). To further dissect the HA module, we analyze the contribution of our Aggregation-Attention. In the w/o Aggre Att variant, we remove only this specific step. The results demonstrate that while this configuration is an improvement over having no HA module at all, it remains significantly inferior to the full CMHANet.
These experiments validate that both the overall iterative architecture of our Hybrid Attention module and the specific design of its cross-modal Aggregation-Attention component are critical to our CMHANet.
The data clearly show that the synergistic integration of these components allows the CMHANet architecture to effectively leverage multimodal information and achieve its superior registration results.

\begin{figure*}[t]
  \centering
  \includegraphics[width=0.95\linewidth]{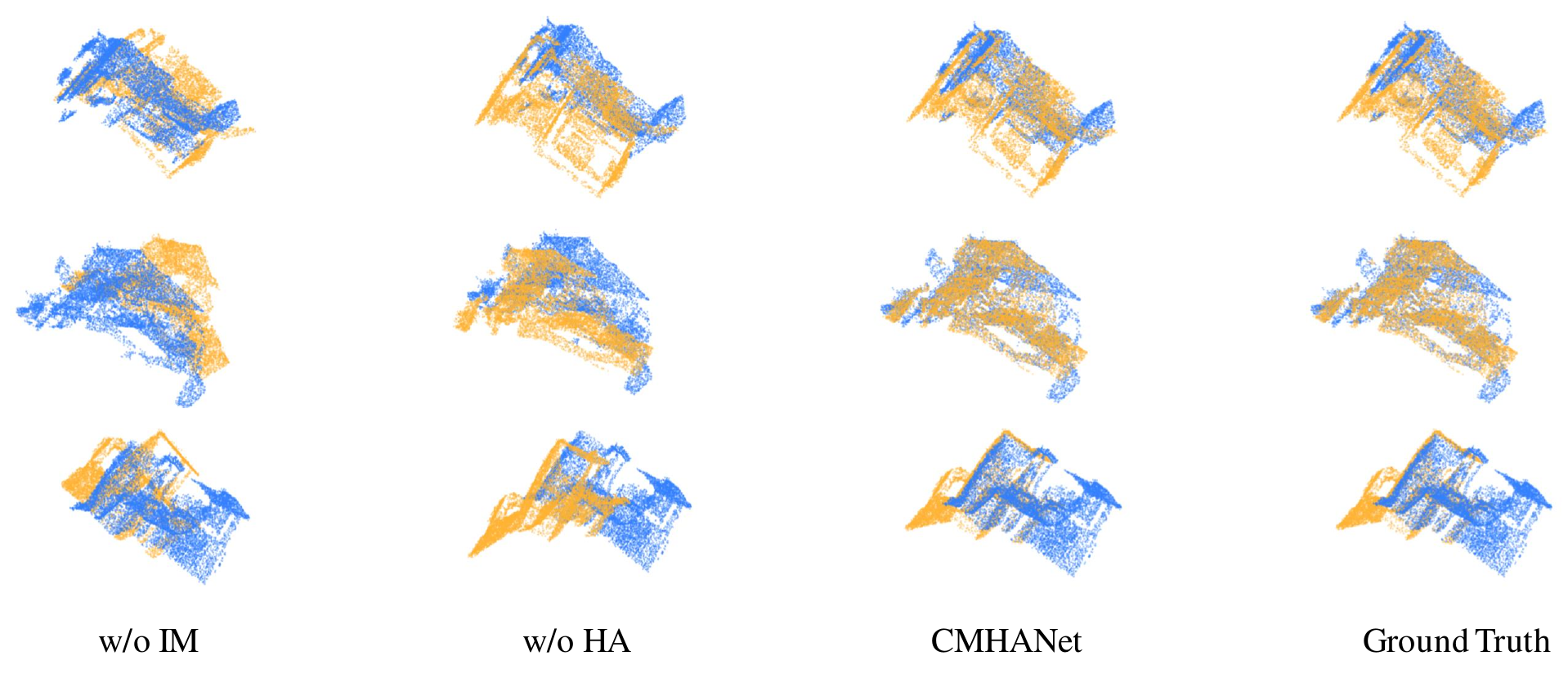}
  \caption{\rev{Visual ablation comparison demonstrating the impact of different feature transformation modules on registration performance. From left to right, w/o IM: Without the Image Module, the network relies solely on geometry; w/o HA: Removing the Hybrid Attention module hinders effective global feature interaction; CMHANet: Our full model leverages cross-modal feature transformation; Ground Truth: Shown for reference.}}
  \label{ablation}
\end{figure*}

\rev{To intuitively understand how our proposed modules handle complex backgrounds, we provide a visual ablation comparison in Figure~\ref{ablation}.
As observed in the first column, removing the Image Module ($w/o$ IM) forces the network to rely solely on geometry. In scenes with repetitive structures or limited geometric overlap, this leads to significant misalignment.
Similarly, the second column ($w/o$ HA) shows that without the global feature transformation provided by Hybrid Attention, the model fails to establish consistent correspondences across the scene.
In contrast, our full CMHANet (Column 3) effectively leverages both visual semantics and attention-based refinement to robustly align the point clouds, closely matching the Ground Truth (Column 4). This confirms that our cross-modal feature transformation is critical for detection performance in challenging scenarios.}

\section{CONCLUSION}
In this work, we propose CMHANet, a point cloud registration method that utilizes a two-stage correspondence process from coarse to fine. The core of our method is the effective fusion of multimodal information, wherein geometric details from 3D point clouds are augmented with contextual features from 2D images via an enhanced attention mechanism. While this cross-modal fusion necessitates a modest increase in inference time and memory footprint compared to single-modal approaches, our experimental results demonstrate that it yields significant improvements in both registration accuracy and robustness. Future work will focus on refining the model for low-overlap conditions, with a specific direction being the decoupling of rotation and translation computations to enhance the alignment process.

\section{Future Work}
Beyond spatial alignment, the cross-modal hybrid attention architecture developed in this work holds significant potential for broader interdisciplinary applications~\citep{wang2023enhancing}.

% \section*{Acknowledgements}
% This work was supported in part by NSFC under grant No. 62125305, the Natural Science Basis Research Plan in Shaanxi Province of China under Grant No. 2025JC-JCQN-091 and the Technology Innovation Leading Program of Shaanxi (Program No. 2024QY-SZX-23).

%% The Appendices part is started with the command \appendix;
%% appendix sections are then done as normal sections

%% \label{}

%% If you have bibdatabase file and want bibtex to generate the
%% bibitems, please use
%%
\bibliographystyle{elsarticle-num-names} 
\bibliography{example}
%% else use the following coding to input the bibitems directly in the
%% TeX file.

%\begin{thebibliography}{00}

%% \bibitem[Author(year)]{label}
%% For example:

%% \bibitem[Aladro et al.(2015)]{Aladro15} Aladro, R., Martín, S., Riquelme, D., et al. 2015, \aas, 579, A101

%\end{thebibliography}

\end{document}